\documentclass{article} 
\usepackage{cite}
\usepackage{amssymb}
\usepackage{amsfonts}
\usepackage{textcomp}
\usepackage{xcolor}
\def\BibTeX{{\rm B\kern-.05em{\sc i\kern-.025em b}\kern-.08em
    T\kern-.1667em\lower.7ex\hbox{E}\kern-.125emX}}
  
\usepackage{changepage}

\usepackage{float}
\usepackage{multirow}
\usepackage{algorithm}
\usepackage{algpseudocode}
\usepackage{pbox}
\algrenewcommand{\algorithmiccomment}[1]{\hfill$\triangleright$\ #1}

\usepackage{dsfont}

\usepackage{tikz}
\usetikzlibrary{positioning}
\usetikzlibrary{shapes.geometric}

\usepackage{ijcai23}

\usepackage{times}
\usepackage{soul}
\usepackage{url}
\usepackage[hidelinks]{hyperref}
\usepackage[utf8]{inputenc}
\usepackage[small]{caption}
\usepackage{graphicx}
\usepackage{amsmath}
\usepackage{amsthm}
\usepackage{booktabs}
\usepackage[switch]{lineno}

\title{Combining Self-labeling with Selective Sampling}

\author{
Jedrzej Kozal$^1$
\and
Michał Woźniak$^1$
\affiliations
$^1$\textit{Department of Systems and Computer Networks} \\
\textit{Wrocław University of Science and Technology}, 
Wrocław, Poland \\
\emails
\{jedrzej.kozal, michal.wozniak\}@pwr.edu.pl
}

\begin{document}



\maketitle

\begin{abstract}


Since data is the fuel that drives machine learning models, and access to labeled data is generally expensive, semi-supervised methods are constantly popular. They enable the acquisition of large datasets without the need for too many expert labels. 
This work combines self-labeling techniques with active learning in a selective sampling scenario. We propose a new method that builds an ensemble classifier. Based on an evaluation of the inconsistency of the decisions of the individual base classifiers for a given observation, a decision is made on whether to request a new label or use the self-labeling.
In preliminary studies, we show that na\"ive application of self-labeling can harm performance by introducing bias towards selected classes and consequently lead to skewed class distribution. Hence, we also propose mechanisms to reduce this phenomenon. 
Experimental evaluation shows that the proposed method matches current selective sampling methods or achieves better results.

\end{abstract}


\section{Introduction}

Active learning \cite{Cohn1994} is the area of machine learning where a training set is constructed by selecting the most informative samples that can speed up training. New labeled learning examples are obtained by queering, i.e., requesting ground truth labels from an oracle. 
To create a query, we use a model trained with a small number of labeled samples.
Stream-Based Selective Sampling \cite{Cohn1994} is based on the assumption that acquiring new unlabeled training examples is relatively inexpensive. We process a single sample at a time, and decide whether it should be labeled by oracle or discarded. 
In this work, we propose a new method that combines self-labeling with active learning in Stream-Based Selective Sampling scenario.

\begin{figure}
    \centering
    \resizebox{200pt}{130pt}{\begin{tikzpicture}[
squarednode/.style={rectangle, draw, minimum size=5mm},
]
\node [squarednode, align=center] (input) {Unlabeled data};

\node[trapezium, minimum width=10mm, draw, rotate=180] (legend1) [left=5.0cm of input] {};
\node[] (legend1_label) [right=1.1cm of legend1] {model};
\node[squarednode, minimum width=7mm,fill={rgb:orange,1;yellow,2;pink,5}] (legend2) [below=0.8cm of legend1] {};
\node[] (legend2_label) [below=0.3cm of legend1_label] {prediction};

\node [] (anchor) [below=0.3cm of input] {};

\node [] (anchor_model2) [left=0.3cm of anchor] {};
\node [] (anchor_model1) [left=1.1cm of anchor_model2] {};
\node [] (anchor_dots) [right=0.8cm of anchor_model2] {};
\node [] (anchor_model3) [right=0.8cm of anchor_dots] {};

\node [trapezium, minimum width=10mm, draw, rotate=180] (model2) [below=0.8cm of anchor_model2] {};
\node [trapezium, minimum width=10mm, draw, rotate=180] (model1) [below=0.8cm of anchor_model1] {};
\node [minimum width=15mm] (dots) [below=0.5cm of anchor_dots] {$\hdots$};
\node [trapezium, minimum width=10mm, draw, rotate=180] (model3) [below=0.8cm of anchor_model3] {};

\node [squarednode, minimum width=7mm,fill={rgb:orange,1;yellow,2;pink,5}] (prediction1) [below=0.8cm of model1] {};
\node [squarednode, minimum width=7mm,fill={rgb:orange,1;yellow,2;pink,5}] (prediction2) [below=0.8cm of model2] {};
\node [minimum width=7mm] (dots_pred) [below=0.5cm of dots] {$\hdots$};
\node [squarednode, minimum width=7mm,fill={rgb:orange,1;yellow,2;pink,5}] (prediction3) [below=0.8cm of model3] {};

\node[] (anchor_out) [below=1.95cm of anchor] {};
\node [] (anchor_out_model2) [below=0.3cm of prediction2] {}; 
\node [] (anchor_out_model1) [below=0.3cm of prediction1] {}; 
\node [] (anchor_out_model3) [below=0.3cm of prediction3] {}; 

\node [squarednode, align=center] (check) [below=0.3cm of anchor_out] {confidence and\\consistency check};

\node [] (failed) [right=0.8cm of check] {FAILED};
\node [] (anchor_if_up) [right=0.8cm of failed] {};
\node [] (passed) [left=0.8cm of check] {PASSED};
\node [] (anchor_if_down) [left=0.8cm of passed] {};

\node [squarednode, align=center, minimum width=17mm] (oracle) [below=0.3cm of anchor_if_up] {request\\label};
\node [squarednode, align=center] (training1) [below=0.3 of oracle] {bootstrapped\\training};
\node [squarednode, align=center, minimum width=17mm] (filter) [below=0.3cm of anchor_if_down] {prior\\filter};
\node [squarednode, align=center] (training2) [below=0.3 of filter] {bootstrapped\\training};

\node[] (ensemble) [right=1.2cm of model3] {ensemble};
\node[] (predictions) [right=0.4cm of prediction3] {predictions};
\node[align=center] (AL) [below=0.2cm of training1] {Training with label from annotator\\(Active Learning)};
\node[align=center] (SL) [below=0.2cm of training2] {Training with predicted label\\(Self-labeling)};

\draw[-] (input.south) -- (anchor.center);
\draw[-] (anchor.center) -- (anchor_model1.center);
\draw[-] (anchor.center) -- (anchor_model3.center);

\draw[->] (anchor_model1.center) -- (model1);
\draw[->] (anchor_model2.center) -- (model2);
\draw[->] (anchor_model3.center) -- (model3);

\draw[->] (model1) -- (prediction1);
\draw[->] (model2) -- (prediction2);
\draw[->] (model3) -- (prediction3);

\draw[-] (prediction1) -- (anchor_out_model1.center);
\draw[-] (prediction2) -- (anchor_out_model2.center);
\draw[-] (prediction3) -- (anchor_out_model3.center);
\draw[-] (anchor_out_model1.center) -- (anchor_out.center);
\draw[-] (anchor_out_model3.center) -- (anchor_out.center);
\draw[->] (anchor_out.center) -- (check);

\draw[-] (check.east) -- (failed.west);
\draw[-] (failed.east) -- (anchor_if_up.center);
\draw[-] (check.west) -- (passed.east);
\draw[-] (passed.west) -- (anchor_if_down.center);

\draw[->] (anchor_if_up.center) -- (oracle.north);
\draw[->] (oracle) -- (training1);
\draw[->] (anchor_if_down.center) -- (filter.north);
\draw[->] (filter) -- (training2);

\end{tikzpicture}}
    \caption{Overview of the proposed method. We utilize ensemble predictions to determine whether a given sample could be added to the dataset with the predicted label (self-labeling) or should be labeled by oracle (active learning). More specifically, we check if obtained support exceeds a predefined threshold and if all confident predictions return to the same class. If not, we check if the budget, create a query, and train with bootstrapping. Otherwise, we filter out and drop the samples from the current majority class (prior filter), and perform bootstrapped training with label obtained from prediction.}
    \label{fig:schema}
\end{figure}
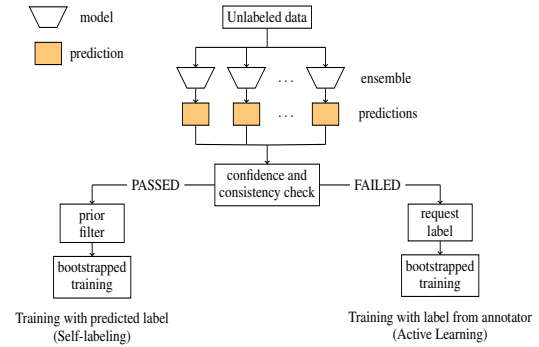

An overview of our method is provided in Fig.~\ref{fig:schema}.
We hope that this approach could allow for the cost-efficient creation of bigger labeled datasets. Self-labeling could introduce noisy labels into the dataset \cite{DBLP:journals/corr/abs-1908-02160}, as in most cases, models have non-zero classification error. In \cite{DBLP:journals/corr/abs-2003-10471} various types of noises and their impact on deep learning performance were analyzed. It was found for various noise types that, with the increase of the noise ratio, test accuracy decreased. and with the increase of dataset size, test accuracy increases. 
In self-labeling, errors made by the classifier introduce wrong labels to the dataset, but as we label new samples, the overall data volume increases. 
If the gain in accuracy from increasing the dataset size surpasses the performance loss from the wrong labels, we can use self-labeling to boost classification performance. In this work, we utilize this dynamic to improve classification performance. 
The main contributions of this work are following:
\begin{itemize}
    \item An analysis of the problems, that arise when we apply self-labeling to an active learning scenario
    \item New method was proposed based on classifier committee, that integrates self-labeling to active learning
    \item Thorough experimental evaluation with multiple dataset and settings
\end{itemize}

\section{Related Works}


\subsection{Active learning}

The most popular active learning strategy is based on uncertainty sampling \cite{10.1007/978-1-4471-2099-5_1}, where model supports are utilized as an information source about learning example usefulness. 
Fragment of the sample space, where the support for the samples is low, is called the region of uncertainty \cite{NIPS1989_b1a59b31}.
This concept was used in \cite{LEWIS1994148} to select samples for labeling with the lowest difference in computed support and the predefined threshold.
In \cite{variable_threshold}, authors proposed uncertainty sampling with a variable threshold for data stream mining. 
In margin sampling \cite{10.1007/3-540-44816-0_31}, queries are created by selecting samples with the smallest difference in probabilities of two classes with the largest confidence. It was shown in \cite{al_experimental_study_baselines} that this algorithm performs on par with more computationally expensive ensemble-based methods. In \cite{DBLP:journals/corr/abs-1906-00025}, modification of standard margin sampling was proposed, 
were samples are selected based on the smallest classification margin of all models in the ensemble.
Query by Committee algorithms measures disagreement between members of an ensemble to choose the most informative samples. In vote entropy, \cite{Argamon_Engelson_1999} samples are selected based on the entropy of ensemble vote distribution. A modified version of this algorithm \cite{juan_s_gomez_canon_2021_5624399} select samples with the highest average predictions entropy. Another possible disagreement measure is maximum disagreement sampling \cite{McCallum1998EmployingEA}, where KL divergence is used. 

\subsection{Self-labeling}

Self-supervised learning \cite{DBLP:journals/corr/abs-2011-00362} aims at learning valuable data representation without annotation.
To obtain good representations, we need to define some pretext tasks for a model to solve.
The first attempts of creating self-supervised learning involved auto-encoders \cite{NIPS2006_5da713a6}, patch location prediction \cite{DBLP:journals/corr/DoerschGE15}, inpainting \cite{DBLP:journals/corr/PathakKDDE16}, or rotation prediction \cite{DBLP:journals/corr/abs-1803-07728}.
Clustering was utilized as a pretext task \cite{DBLP:journals/corr/abs-1807-05520,DBLP:journals/corr/abs-1911-05371} for training deep data representations. 
Another research direction is contrastive learning \cite{DBLP:journals/corr/abs-2002-05709}, where the pretext task is based on learning close representations for the same sample with different augmentations applied and distant representations for dissimilar samples.
Pseudo-labels are also used for semi-supervised learning \cite{DBLP:journals/corr/abs-2101-06329}. Authors of \cite{Lee2013PseudoLabelT} utilize the outputs of the classifier with the highest confidence as a target for unlabeled data. In \cite{DBLP:journals/corr/abs-2001-07685}, new methods were proposed that utilize high-confidence pseudo-labels generated with weakly-augmented images. Next, these pseudo-labels are used as a target for the strongly-augmented version of the same image. 

\subsection{Active learning with Self-labeling}

In \cite{DBLP:journals/corr/WangZLZL17} combination of automatic pseudo-labeling and active learning was proposed for a pool-based setting. Authors found that utilization of pseudo-labels can improve the labeling efficiency of active learning algorithms, and error rates of automatically assigned labels are low for the convolutional neural network.
Authors of \cite{DBLP:journals/corr/abs-1911-08177} introduce a new method that combines semi-supervised learning with pseudo-labels and active learning.
Korycki and Krawczyk \cite{DBLP:journals/corr/abs-2112-11019} have combined self-labeling with active learning for learning from data streams.

\section{Method}

In this section, we describe the setting and introduce our method. We also provide results of preliminary experiments with the dynamic imbalance.

\subsection{Selective sampling}

First, we introduce the selective sampling framework that our work is based on. We assume an access to small set of labeled data $\mathcal{L} = \{ (\boldsymbol{x}_m, y_m) \} _{m=1}^M$ and stream of unlabeled data $\mathcal{U} = \{ (\boldsymbol{x}_n \} _{n=1}^N$ with $x \in \mathcal{X}, y \in \mathcal{Y}$, where $\mathcal{X}$ and $\mathcal{Y}$ are the input space and the set of labels respectively. Our goal is to train a model $f$, that predicts the support for input sample $\boldsymbol{x}$, namely: $p(\boldsymbol{y}|\boldsymbol{x}) = f_{\theta}(\boldsymbol{x})$, where $\theta$ is set of model parameters. Final model prediction is given by $\hat{y} = arg\max_{i} p(\boldsymbol{y}_{i}|\boldsymbol{x})$.
We denote maximum support for sample $\boldsymbol{x}$ as $\max_{i} p(\boldsymbol{y}_i|\boldsymbol{x})$.
The general algorithm for selective sampling is provided in appendix~\ref{appendix:algorithm_selective_sampling}. 
We assume the same cost of obtaining label from an oracle for each sample in $\mathcal{U}$. For this reason, we define budget $B$ as the number of samples that can be labeled, with the exception of presenting results when we refer to budget as the fraction of all samples that can be labeled.

\subsection{Informativeness computation}

We employ differences in supports obtained from different models in an ensemble as a source of informativeness. First, the ensemble of $L$ base classifiers is trained. For the unlabeled sample $\boldsymbol{x}$ each model $l$ in a committee computes supports $p_{l}(\boldsymbol{y}|\boldsymbol{x})$. Next, we check if at least half of the classifiers in the ensemble provided supports that exceed a predefined support threshold $\tau$. 

\begin{equation}
    \sum_{l} \mathds{1}_{\max_{c} p_{l}(\boldsymbol{y}_c|\boldsymbol{x}) > \tau} > \frac{L}{2}
\end{equation}

If more than half of the models return confident predictions and these models output the same prediction, we add $(\boldsymbol{x}, \hat{y})$ to $\mathcal{L}$. Otherwise, we query an oracle with $\boldsymbol{x}$. By choosing samples with consistent, highly confident predictions, we avoid assigning the wrong label to a sample.
From an active learning perspective, these learning examples are not valuable, as models already return confident predictions for them. However, we hope that by a faster increase in the number of labeled samples, we can obtain improvements in classification accuracy.

\subsection{Bootstrapped training}

We train initial models with bootstrapping of labeled part of data $\mathcal{L}$.
This corresponds to sampling number of repeats of each sample from the Poisson distribution with $\lambda = 1$. This part of our method is inspired by Online Bagging \cite{Oza:2001} method, introduced for data stream classification. During training with an unlabeled stream, we use bootstrapping for new learning examples added to the dataset. In the case of training with ground truth label from oracle, we use $\lambda = 1$. When updating the dataset with a sample labeled based on model prediction, we calculate $\lambda$ as:

\begin{equation}
    \lambda = \frac{ \max_{l, c} p_l(\boldsymbol{y}_c|\boldsymbol{x}) }{\tau} - \mathds{1}_{B = 0}
    \label{eq:lambda}
\end{equation}
where $\tau$ is the same threshold used earlier for selecting confident predictions. 
When $B > 0$, then $\lambda$ is always greater than one.
As a result, samples labeled based on model prediction will be more frequently added to the dataset than learning examples from initial dataset $\mathcal{L}$. 
To avoid the negative influence of incorrect model predictions after the budget ended we change lambda calculation after labeling budget was spent. In such case $\lambda < 1$, 
assuming that value of $\tau$ is not significantly lower than $p$. Consequently, updates to datasets are still performed, while the negative impact of incorrect predictions is limited. Values of $\lambda$ for each sample are stored in $ \boldsymbol{\lambda}$ vector. Upon an update, we generate separate datasets by bootstrapping. The number of repeats of a single sample in a dataset is limited to 4. The ensemble training procedure for the proposed method is given in algorithm \ref{alg:ensemble_training}. 

\begin{algorithm}
\caption{Bootstrapped training}
\begin{algorithmic}[1]
\Require $\mathcal{L}$ - set of labeled data with $M$ elements, $ \{ f_{\theta} \}_{L} $ - ensemble of $L$ models, $\boldsymbol{\lambda}$ - vector with parameters for Poisson distribution for each sample in $\mathcal{L}$

\For{$l \in \{0, L\}$}
    \State $\boldsymbol{r} \thicksim Pois(\boldsymbol{\lambda})$
    \State $\boldsymbol{r} \leftarrow \min (\boldsymbol{r}, 4)$
    \State $D \leftarrow \emptyset$
    \For{$i \in \{0, M\}$}
        \State $(\boldsymbol{x}, y) \leftarrow \mathcal{L}_{i}$
        \For{$j \in \{0, \boldsymbol{r}_{i}\}$}
            \State $D \leftarrow D \cup \{(\boldsymbol{x}, y)\} $
        \EndFor
    \EndFor
    \State train $f_{\theta_{l}}$ with $D$
\EndFor
\end{algorithmic}
\label{alg:ensemble_training}
\end{algorithm}

\subsection{Dynamic Imbalance}

Na\"ive usage of self-labeling in selective sampling can introduce imbalance in the training set $\mathcal{L}$. To demonstrate this, we conduct a preliminary experiment with synthetic data. We generate simple datasets by sampling from 2D Gaussian distribution for easier visualization. We study two scenarios that may occur in practice. In the first case, the dataset consists of three balanced classes, but one of the classes is easier to learn than the rest. In the second scenario dataset with two classes is imbalanced.

\begin{figure*}
    \centering
    \includegraphics[width=0.8\textwidth]{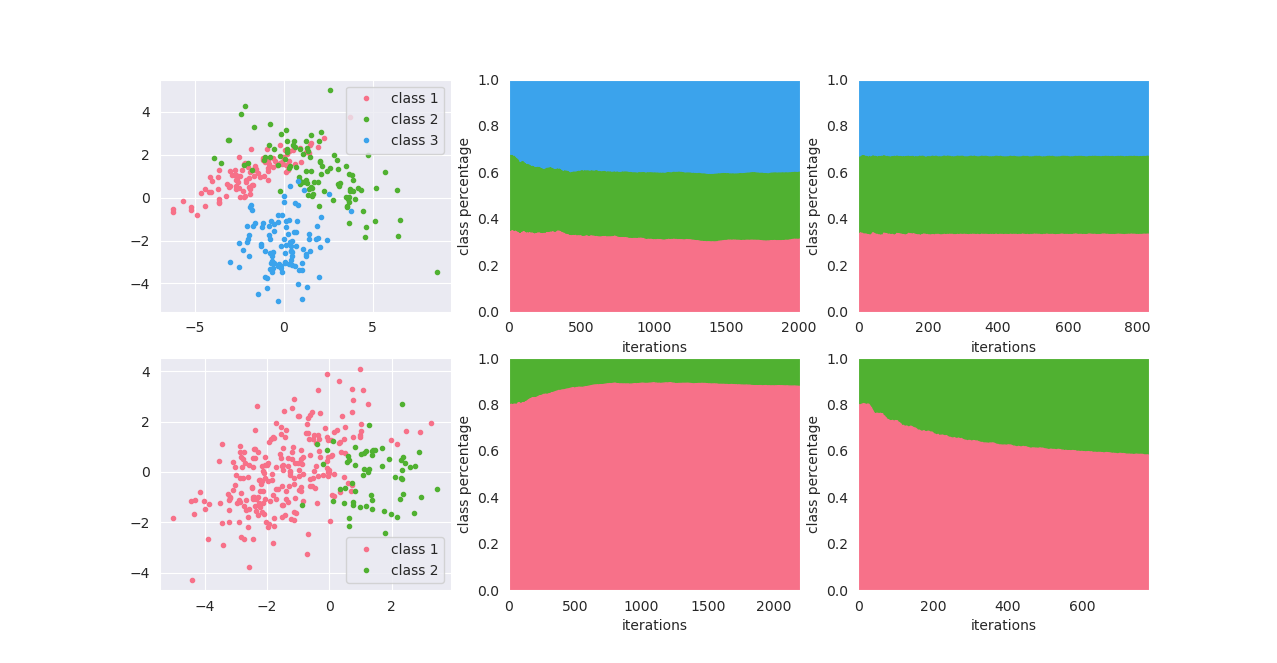}
    \caption{Dynamic imbalance of classes when applying self-labeling directly during selective sampling. We consider two settings: in the first three classes with balanced prior distribution, but a single class is easier to learn than others (up), and the second imbalanced binary classification problem (bottom). Generated 2-D datasets are plotted on the left-hand side. When applying self-labeling directly (middle), we observe a change in the class distribution in the training set. This problem can be avoided when we apply dynamic balancing (right).}
    \label{fig:imbalance}
\end{figure*}

We sample 300 learning examples, plotted in Fig.~\ref{fig:imbalance} on the left-hand side. Datasets are used for initial training of Multi-layer Perceptron with 5 neuron single hidden layer. Next, we generate a stream with 3000 learning examples, sample data from the stream, and obtain model predictions. When model confidence exceeds 0.95, we expand the training set with learning examples and predicted labels. For simplicity, we do not use bootstrapping in this experiment. When model confidence is below 0.7, we create a query to obtain a ground-truth label. Changes in the percentage of labels in training set during training with an unlabeled data stream are presented in Fig.~\ref{fig:imbalance} in the middle.

In the first scenario, over time percentage of samples labeled as the third class grows until it utilizes approximately 40\% of all data. This result shows that na\"ive utilization of self-labeling can disturb class distribution, even if the original data is balanced. In the second scenario, the initial imbalance ratio is 1:4, however, after approximately 800 iterations, it is closer to 1:5. This shows that initial bias in the data distribution can be strengthened by self-labeling. Please note that in this experiment algorithm have access to the ground truth labels by creating queries for samples with low confidence, and yet, the class distribution change over time. 

\subsection{Prior filter}

To address the issue of dynamic imbalance, we introduce a method that prevents training when the current prior estimation for the predicted class is too high. We use the last $k$ labels from $\mathcal{L}$ and compute the percentage of samples that have the same label as the predicted class:

\begin{equation}
    \hat{p} = \frac{1}{k} \sum_{i=M-k}^{M} \mathds{1}_{y_{i}=\hat{y}}
\end{equation}

This value can be interpreted as an estimation of the current class prior.
Only the last $k$ labels were used because, as shown in preliminary experiments, class distribution can change over time.
We compute difference between $\hat{p}$ and prior of perfectly balanced dataset:

\begin{equation}
    \Delta_{p} = \hat{p} - \frac{1}{C}
\end{equation}

Where $C$ is the number of all classes. When $\Delta_{p} > 0$ we disallow training. We do not apply this prior filter to labels obtained from an oracle. 
A similar approach was proposed earlier in \cite{9892104} in the context of the data stream processing, however it estimated prior with regression models and switching labels for the majority class.
Here we estimate prior directly from model predictions and skip samples from majority classes, which is similar to undersampling.

We repeat previous preliminary experiments with the prior filter applied. We use $k=50$ last samples. Results are plotted in Fig.~\ref{fig:imbalance} on the right-hand side. The proposed method can keep class distribution balanced in the first setting and, over time, improve initial class distribution in the second setting. 

\subsection{Self-labeling selective sampling}

The complete algorithm for Self-labeling selective sampling (SL2S) along with time complexity analysis is provided in appendix \ref{appendix:detailed_algorithm}.

\section{Experimental Setup}

This section provides a detailed description of the methods, datasets, and tools used to conduct experiments.

\subsection{Datasets}

We utilize datasets from the UCI repository \cite{uci} with a wide range of datasets with different sizes, number of classes, number of attributes, and imbalance ratio (IR). The detailed information about data used in experiments is presented in Tab.~\ref{tab:datasets}.
The complete list of features and procedures for loading data are provided in appendix \ref{appendix:data_loading}.

\begin{table}[ht]
    \scriptsize
    \centering
    \caption{Datasets used for experiments. IR was computed by taking a ratio of class with the highest and lowest number of samples.}
    \begin{tabular}{l|c|c|c|c}
    \hline
    dataset name & size & \#class & \#attributes & IR \\
    \hline
 adult \cite{adult}                            & 48842    & 2    & 14           & 3.1527       \\
 bank marketing \cite{bank}                    & 45211    & 2    & 17           & 7.5475       \\
 firewall \cite{firewall}                      & 65478    & 3    & 12           & 2.9290       \\
 chess \cite{uci} & 20902 & 15 & 40 & 22.919 \\
    \hline
 nursery \cite{nursery}                        & 12958    & 4    & 8            & 13.1707      \\
 mushroom \cite{uci}                           & 8124     & 2    & 22           & 1.0746       \\
 wine \cite{wine}                              & 4873     & 5    & 12           & 13.5082      \\
 abalone \cite{abalone}                        & 4098     & 11    & 8            & 21.5417      \\
    \hline
    \end{tabular}
    \label{tab:datasets}
\end{table}

\subsection{Metrics and evaluation}

Due to high values of IR for some datasets, we decided to employ balanced accuracy \cite{5597285} as primary performance metrics for our experiments. In our evaluation, we focus on the impact of budget size and seed size used for training of the initial model, as these two factors can impact the results the most.
All values of metrics reported in this paper were obtained with a separate test set.
Code was implemented in Python with the utilization of scikit-learn library \cite{scikit-learn}. The codebase with the method and experiment implementations are available on github \footnote{https://github.com/w4k2/active-learning-data-streams}.

\subsection{Baselines}

To perform fair evaluation, we compare the proposed method to commonly used algorithms in selective sampling literature:

\begin{itemize}
    \item random - random selection of samples for query
    \item fixed uncertainty \cite{LEWIS1994148} - selection of samples based on a static confidence threshold
    \item variable uncertainty \cite{variable_threshold} - modification of fixed uncertainty that adjust confidence threshold based on the current size of the uncertainty region
    \item classification margin \cite{10.1007/3-540-44816-0_31} - a method that computes the difference in confidence between classes with two biggest supports
    \item vote entropy \cite{Argamon_Engelson_1999} - queries are based on ensemble vote entropy
    \item consensus entropy \cite{juan_s_gomez_canon_2021_5624399} - samples are selected based on the highest average prediction entropy
    \item max disagreement \cite{McCallum1998EmployingEA} - computes KL-divergence between output class distribution and consensus distribution
    \item min margin \cite{DBLP:journals/corr/abs-1906-00025} - a method that selects samples based on minimum classification margin for all models in the ensemble
\end{itemize}

In the case of methods that were created for the pool-based scenario, we adapt them by introducing the informativeness threshold.
For each committee-based method, we use 9 base classifiers and employ bootstrapping during initial training. All methods were trained with Multi-layer Perceptron classifier with two hidden layers, 100 neurons each.

\subsection{Hyperparameter tuning}

In preliminary experiments, we found that the most important hyperparameter is the threshold used for the informativeness measure. For this reason, we focused on tuning this parameter. We use random search \cite{JMLR:v13:bergstra12a} to select the best thresholds for each algorithm. MLP classifier was trained with Adam optimizer (learning rate equal to 0.001). We allow training for a maximum of 5000 iterations. Detailed description of hyperparameter tuning process with range of values for each algorithm are provided in appendix~\ref{appendix:hyperparamters}.

\subsection{Goal of experiments}

The overall goal of experiments is to perform a thorough investigation into the usefulness of self-labeling in a selective sampling setting.
To provide a more precise description, we formulate the following research questions:

\begin{itemize}
    \item[RQ1:]Is there a benefit of combining active learning strategies with self-labeling?
    \item[RQ2:]What is the performance of the proposed method for datasets with a high number of learning examples?
    \item[RQ3:]What is the impact of the initial training size (the seed size) on the performance?
    \item[RQ4:]How does the accuracy of the model trained with seed impact the learning process of the proposed algorithm?
    \item[RQ5:]Does the proposed algorithm allows for the better utilization of the computational budget?
\end{itemize}

Each of these research questions will be addressed in the following parts of our work.

\section{Experiments}

In this section, we describe the results of an experimental evaluation in accordance with the research questions stated above.

\subsection{Experiments with smaller datasets}

We compare the performance of the proposed method and baselines according to the experimental protocol described in previous sections. Here we utilize four datasets, namely nursery, mushroom, wine, and abalone. Results are presented on the left-hand side of Tab.~\ref{tab:variable_budget}. 

\begin{table*}[!ht]
    \scriptsize
    \centering
    \caption{Balanced accuracy for variable budget size and smaller datasets}
    \addtolength{\tabcolsep}{-4pt} 
    \begin{tabular}{l|ccccc|ccccc}

\hline
dataset & \multicolumn{5}{c}{nursery} & \multicolumn{5}{c}{adult} \\ 
\hline
 labeled           & & & 0.318±0.030 & &  & & & 0.741±0.010 & & \\
 labeled ensemble   & & &  0.276±0.013  & &   & & &  0.756±0.005  & & \\
\hline
budget & 0.1 & 0.2 & 0.3 & 0.4 & 0.5 & 0.1 & 0.2 & 0.3 & 0.4 & 0.5 \\ 
\hline
 random                 &  0.371±0.015  &  0.350±0.017  &  0.325±0.012  &  0.298±0.017  &  0.282±0.012 & 0.735±0.007  &  0.733±0.005  &  0.729±0.007  &  0.731±0.005  &  0.732±0.006 \\
 f. uncertainty      &  0.389±0.018  &  0.393±0.016  &  0.385±0.007  &  0.391±0.016  &  0.394±0.019 & 0.754±0.007  &  0.758±0.008  &  0.765±0.009  &  0.760±0.011  &  0.760±0.011 \\
 v. uncertainty   &  0.378±0.012  &  0.359±0.012  &  0.327±0.014  &  0.307±0.014  &  0.280±0.018 & 0.756±0.013  &  0.751±0.011  &  0.755±0.012  &  0.758±0.012  &  0.746±0.011 \\
 class. margin  &  0.397±0.019  & \textbf{  0.395±0.020  } & \textbf{  0.396±0.013  } &  0.399±0.036  &  0.396±0.019 & 0.757±0.008  &  0.757±0.008  &  0.757±0.008  &  0.757±0.008  &  0.757±0.008 \\
 vote entropy           &  0.393±0.014  &  0.393±0.014  &  0.393±0.014  &  0.393±0.014  &  0.393±0.014 & - & - & - & - & - \\
 consensus entropy      &  0.393±0.013  &  0.394±0.013  &  0.393±0.014  &  0.393±0.013  & \textbf{  0.404±0.017 } & 0.764±0.004  & \textbf{  0.767±0.005  } &  0.765±0.002  &  0.765±0.004  &  0.764±0.003 \\
 max disagreement       &  0.402±0.019  &  0.393±0.014  &  0.393±0.014  & \textbf{  0.404±0.016  } &  0.403±0.021  & - & - & - & - & - \\
 min margin             & \textbf{  0.405±0.019  } &  0.375±0.012  &  0.388±0.021  &  0.385±0.018  &  0.400±0.010 & \textbf{  0.768±0.004  } & \textbf{  0.767±0.005  } & \textbf{  0.768±0.004  } & \textbf{  0.768±0.004  } & \textbf{  0.768±0.004 } \\
 SL2S                   &  0.384±0.018  &  0.350±0.014  &  0.338±0.013  &  0.292±0.020  &  0.294±0.016 & 0.762±0.003  &  0.763±0.004  &  0.763±0.004  &  0.762±0.003  &  0.762±0.004 \\

\hline
dataset & \multicolumn{5}{c}{mushroom} & \multicolumn{5}{c}{bank marketing} \\ 
\hline
 labeled           & & & 0.637±0.010 & &   & & & 0.712±0.012 & & \\
 labeled ensemble   & & &  0.636±0.010  & &    & & & 0.714±0.009 & & \\
\hline
budget & 0.1 & 0.2 & 0.3 & 0.4 & 0.5 & 0.1 & 0.2 & 0.3 & 0.4 & 0.5 \\ 
\hline
 random                 & \textbf{  0.632±0.011  } & \textbf{  0.634±0.009  } &  0.633±0.010  & \textbf{  0.636±0.010  } &  0.633±0.012 
 &  0.700±0.023  &  0.694±0.017  &  0.703±0.021  &  0.698±0.014  &  0.699±0.018 \\
 f. uncertainty      &  0.630±0.010  &  0.630±0.011  &  0.633±0.012  &  0.635±0.009  &  0.634±0.010 
 &  0.691±0.014  &  0.710±0.014  &  0.705±0.019  &  0.705±0.019  &  0.705±0.019 \\
 v. uncertainty   &  0.631±0.010  &  0.631±0.012  &  0.634±0.010  &  0.633±0.009  &  0.633±0.011 &  0.690±0.014  &  0.694±0.018  &  0.700±0.018  &  0.703±0.013  &  0.697±0.018 \\
 class. margin  & \textbf{  0.632±0.012  } &  0.633±0.012  &  0.633±0.013  &  0.618±0.023  & \textbf{  0.635±0.010 } &  0.682±0.012  &  0.682±0.012  &  0.682±0.012  &  0.682±0.012  &  0.682±0.012 \\
 vote entropy           &  0.630±0.011  &  0.630±0.011  & \textbf{  0.635±0.011  } &  0.630±0.011  & \textbf{  0.635±0.010 } 
 & - & - & - & - & - \\
 consensus entropy      & \textbf{  0.632±0.010  } &  0.632±0.012  &  0.633±0.011  &  0.631±0.011  &  0.634±0.010 &  0.701±0.011  &  0.714±0.007  & \textbf{  0.719±0.006  } &  0.715±0.009  &  0.715±0.009 \\
 max disagreement       &  0.631±0.010  &  0.633±0.013  &  0.630±0.011  &  0.630±0.011  &  0.630±0.011 
 & - & - & - & - & - \\
 min margin             & \textbf{  0.632±0.011  } &  0.632±0.012  &  0.634±0.010  &  0.634±0.011  &  0.634±0.011 
 &  0.705±0.006  & \textbf{  0.716±0.007  } &  0.716±0.007  &  0.716±0.007  &  0.716±0.007 \\
 SL2S                   &  0.631±0.011  &  0.632±0.012  &  0.632±0.012  &  0.633±0.010  &  0.634±0.010 & \textbf{  0.709±0.007  } &  0.715±0.009  &  0.718±0.009  & \textbf{  0.719±0.008  } & \textbf{  0.718±0.009 } \\

\hline 
dataset & \multicolumn{5}{c}{wine} & \multicolumn{5}{c}{firewall} \\ 
\hline 
 labeled           & & & 0.524±0.027 & &    & & & 0.997±0.001 & &  \\
 labeled ensemble   & & &  0.514±0.015  & &     & & & 0.998±0.000 & & \\
\hline
budget & 0.1 & 0.2 & 0.3 & 0.4 & 0.5 & 0.1 & 0.2 & 0.3 & 0.4 & 0.5 \\ 
\hline 
 random                 &  0.408±0.021  &  0.430±0.021  &  0.439±0.023  &  0.452±0.018  &  0.474±0.017 
 &  0.996±0.002  & \textbf{  0.997±0.002  } & \textbf{  0.998±0.001  } & \textbf{  0.997±0.001  } & \textbf{  0.998±0.001 } \\
 f. uncertainty      &  0.418±0.020  &  0.423±0.018  &  0.441±0.017  &  0.448±0.012  &  0.440±0.012 
 &  0.993±0.002  &  0.996±0.002  &  0.996±0.002  &  0.996±0.002  &  0.996±0.002 \\
 v. uncertainty   &  0.415±0.022  &  0.437±0.016  &  0.437±0.022  &  0.459±0.021  &  0.473±0.022 &  0.994±0.002  & \textbf{  0.997±0.001  } &  0.997±0.001  & \textbf{  0.997±0.001  } &  0.997±0.001 \\
 class. margin  &  0.414±0.015  &  0.433±0.016  &  0.420±0.022  &  0.424±0.014  &  0.429±0.021 
 &  0.991±0.001  &  0.991±0.001  &  0.991±0.001  &  0.991±0.001  &  0.991±0.001 \\
 vote entropy           &  0.404±0.012  &  0.419±0.016  &  0.389±0.016  &  0.389±0.016  &  0.389±0.016 
 & - & - & - & - & - \\
 consensus entropy      &  0.419±0.010  & \textbf{  0.439±0.014  } &  0.458±0.014  &  0.474±0.012  &  0.479±0.013 
 &  0.992±0.001  &  0.992±0.001  &  0.992±0.001  &  0.992±0.001  &  0.992±0.001 \\
 max disagreement       &  0.396±0.022  &  0.389±0.016  &  0.389±0.016  &  0.389±0.016  &  0.389±0.016 
 & - & - & - & - & - \\
 min margin             & \textbf{  0.420±0.014  } &  0.432±0.017  & \textbf{  0.461±0.019  } &  0.465±0.016  &  0.487±0.012 
 &  0.991±0.001  &  0.991±0.001  &  0.991±0.001  &  0.991±0.001  &  0.991±0.001 \\
 SL2S                   & \textbf{  0.420±0.014  } &  0.438±0.009  &  0.451±0.022  & \textbf{  0.476±0.014  } & \textbf{  0.499±0.019 } 
 & \textbf{  0.997±0.001  } & \textbf{  0.997±0.001  } &  0.997±0.001  & \textbf{  0.997±0.001  } &  0.997±0.001 \\

\hline 
dataset & \multicolumn{5}{c}{abalone} & \multicolumn{5}{c}{chess} \\ 
\hline 
 labeled           & & & 0.186±0.021 & &       & & & 0.816±0.011 & & \\
 labeled ensemble   & & &  0.188±0.012  & &    & & & 0.850±0.007 & & \\
\hline 
budget & 0.1 & 0.2 & 0.3 & 0.4 & 0.5 & 0.1 & 0.2 & 0.3 & 0.4 & 0.5 \\ 
\hline 
 random                 &  0.179±0.012  &  0.177±0.009  &  0.178±0.007  &  0.185±0.011  &  0.182±0.018 
 & \textbf{  0.515±0.017  } &  0.581±0.015  &  0.630±0.009  &  0.669±0.014  &  0.698±0.009 \\
 f. uncertainty      &  0.171±0.024  &  0.187±0.008  &  0.177±0.013  &  0.182±0.014  &  0.186±0.018 
 &  0.459±0.017  &  0.538±0.015  &  0.593±0.017  &  0.639±0.017  &  0.674±0.014 \\
 v. uncertainty   &  0.171±0.012  &  0.182±0.014  &  0.181±0.010  &  0.176±0.006  &  0.181±0.015 
 &  0.464±0.021  &  0.543±0.016  &  0.604±0.016  &  0.651±0.016  &  0.691±0.014 \\
 class. margin  &  0.176±0.016  &  0.181±0.008  &  0.187±0.013  &  0.177±0.018  &  0.183±0.010 
 &  0.466±0.019  &  0.543±0.015  &  0.606±0.013  &  0.653±0.014  &  0.692±0.014 \\
 vote entropy           &  0.185±0.015  &  0.187±0.012  &  0.185±0.015  &  0.185±0.015  & \textbf{  0.188±0.012 } 
 & - & - & - & - & - \\
 consensus entropy      & \textbf{  0.188±0.015  } & \textbf{  0.191±0.010  } &  0.185±0.013  &  0.184±0.017  &  0.187±0.014 
 &  0.492±0.016  & \textbf{  0.594±0.009  } & \textbf{  0.662±0.008  } & \textbf{  0.715±0.010  } & \textbf{  0.758±0.009 } \\
 max disagreement       &  0.183±0.013  &  0.185±0.012  &  0.184±0.012  &  0.185±0.015  &  0.185±0.011 
 & - & - & - & - & - \\
 min margin             &  0.184±0.009  &  0.189±0.012  &  0.185±0.015  & \textbf{  0.188±0.014  } &  0.184±0.014 
 &  0.491±0.022  &  0.589±0.017  &  0.660±0.016  &  0.713±0.016  &  0.752±0.014 \\
 SL2S                   &  0.183±0.013  &  0.186±0.010  & \textbf{  0.190±0.010  } & \textbf{  0.188±0.008  } &  0.182±0.011 &  0.491±0.013  &  0.585±0.010  &  0.655±0.012  &  0.702±0.010  &  0.748±0.010 \\
\hline
\end{tabular}
    \addtolength{\tabcolsep}{4pt} 
    \label{tab:variable_budget}
\end{table*}

Here we can see that our method rarely obtains the best score. However, the difference between the best-performing method and SL2S is often close to or below 0.02. The worst performance is obtained for the nursery dataset. This is probably due to the presence of three majority classes with a close number of samples and a single minority class with a substantially lower number of samples. For this reason, a lot of samples could be discarded by the prior filter. 
Other datasets are either well-balanced or contain a single majority class. For these types of datasets, we obtained better results.

When we compare the performance of other methods, we can notice that uncertainty-based methods perform comparably to the best algorithms only with a high budget. Classification margin is a strong baseline as indicated by literature \cite{https://doi.org/10.48550/arxiv.2210.03822}. When we compare ensemble-based methods, it turns out that min margin and consensus entropy are the best, with both methods obtaining close performance to the best algorithm. 

\subsection{Experiments with bigger datasets}

We also conduct experiments on larger datasets, i.e., adult, bank marketing, firewall, and chess.
To save computation, we train only when batch of 100 labeled samples is collected and reuse the hyperparameters found for the biggest datasets in previous experiments.
We also drop the vote entropy and max disagreement methods from our comparison due to poor performance in previous experiments compared to other ensemble-based methods.
Results are presented on the right-hand side of Tab.~\ref{tab:variable_budget}.

Here our method performs well, with either the best-balanced accuracy or close to the best. There is no clear performance pattern when we compare results across the varying budget. Uncertainty-based methods provide the worst balanced accuracy in this case. Random sampling allows for obtaining the best performance for the firewall dataset, probably due to the simplicity of the classification problem in this dataset. Firewall has only three classes and a lower IR compared to other datasets. Most methods perform well on this dataset, with a lot of ties between different algorithms in the first place. 

\subsection{Impact of seed size}

We evaluate the impact of the size of the initial training set on active learning performance. When utilizing labels generated with model predictions, the lower number of initial training samples may cause a higher error rate at the beginning of the experiment and the introduction of more noise into the dataset. For this reason, smaller seed sizes can impact the overall results. We reuse the hyperparameter values from previous experiments. All experiments are performed with a budget equal to 0.3. The results are provided in appendix \ref{appendix:seed_size}.

As expected, initial training size has a lower impact on the random sampling algorithm. This method is not dependent on model predictions, therefore, changing the seed should not impact the overall performance. In the case of uncertainty-based methods, there is no clear pattern of seed size impact. In some cases, training with these algorithms and lower seed could provide better results. The ensemble-based methods improve balanced accuracy as the number of labeled samples grows. SL2S can in some cases, obtain better performance with smaller seeds, and often we were able to obtain the best-balanced accuracy with our method. This result indicates that SL2S does not depend heavily on the initial model performance and could be applied even if the number of labeled samples in the beginning is small.

\subsection{Ablation studies}

We perform ablation studies for the proposed method. First, we remove the prior filter and allow training regardless of the current dataset imbalance. This modification should further verify whether the dynamic imbalance is an issue when we use self-labeling in selective sampling. Secondly, we keep higher lambda values in equation~\ref{eq:lambda} after the end of the budget. Decreasing lambda is the second mechanism introduced in our work that, in principle, should prevent the gradual degradation of model performance when using self-supervision as a source of new labels. Next, we remove the bootstrapped training to evaluate if ensemble diversification provides better performance in our experiments. Lastly, the self-labeling part of our approach was removed, and training was conducted with active learning alone.
We use the wine dataset for evaluation. Experiments were performed with a 0.3 labeling budget and various seed sizes. 
The prediction threshold value was selected based on hyperparameter tuning results from previous experiments. We repeat experiments with three different random seeds and report average results in Tab.~\ref{tab:ablations}.


\begin{table}[]
    \scriptsize
    \centering
    \caption{Balanced accuracy obtained in ablation study}
    \begin{tabular}{c|c|c|c}
    \hline
        seed size & 100 & 500 & 1000  \\
        \hline
        base & 0.4151±0.0220 & 0.4323±0.0230 & 0.4509±0.0131 \\
        -prior filter & 0.3747±0.0192 & 0.4430±0.0193 & 0.4502±0.0231  \\
        -lambda reduction & 0.3747±0.0192 & 0.4257±0.0159 & 0.4463±0.0226 \\
        -self-labeling & 0.4174±0.0168 & 0.4421±0.0292 & 0.4461±0.0398 \\
        -bootstrapped training & 0.3916±0.0211 & 0.4318±0.0217 & 0.4461±0.0135 \\
        \hline
    \end{tabular}
    \label{tab:ablations}
\end{table}

We find that the prior filter has only a positive impact only in the case of smaller seed sizes. Conversely, reducing lambda after budget end provides gains in balanced accuracy for higher seed size. Removing self-labeling increase accuracy. This finding is expected, as in preliminary experiments we found that na\"ive application of self-labeling could make results worse. In this case, after removing two mechanisms from our algorithm that prevent the negative impact of dynamic imbalance and classification errors, we can see that further removing self-labeling improves the results. This is in line with preliminary results and further proves that prior filter and lambda reduction are indeed necessary. Lastly, the removal of bootstrapped training has a bigger impact when training with a smaller seed size. We can intuitively explain this result by the fact that ensemble diversity should be smaller when utilizing bigger datasets, as more samples could cover feature space more densely, and randomly sampling datasets with bootstrapping would produce more similar datasets.

\subsection{Incorrect labels from self-labeling}

\begin{figure}
    \centering
    \includegraphics[width=0.4\textwidth]{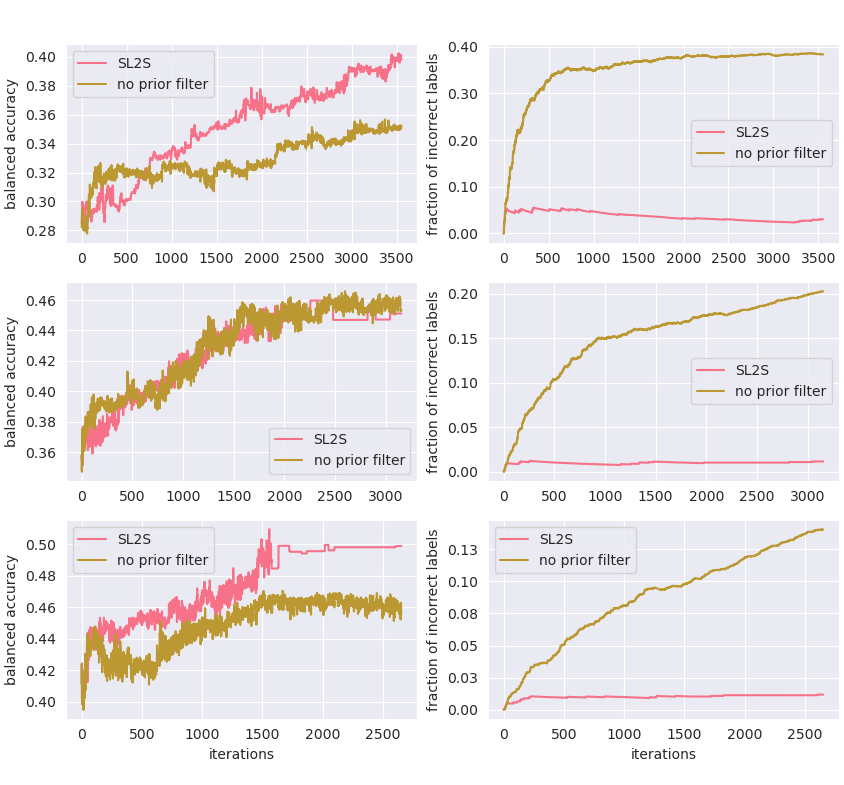}
    \caption{Balanced accuracy (left) with a corresponding fraction of incorrect samples in the labeled dataset (right) over multiple iterations. We perform experiments for various seed sizes: 100 (top), 500 (middle), and 1000 (bottom).}
    \label{fig:frac_incrorect}
\end{figure}

As an extension of ablation studies, we examine how many wrong labels are introduced when using SL2S with and without prior filter. For this purpose, we train the MLP model on a wine dataset with a budget of 0.3 and various seed sizes. We plot the balanced accuracy with the corresponding fraction of samples with wrong labels in the training dataset over multiple iterations in Fig.~\ref{fig:frac_incrorect}.
Including a prior filter drastically reduces the number of incorrect labels. This does not necessarily lead to improvement in balanced accuracy. For a seed size equal to 500, both versions of the algorithm obtain very close final accuracy, while the difference in a fraction of incorrect labels is nearly 0.2. This phenomenon can be explained by two factors. First is the fact that neural networks trained with gradient descent are known to be robust to noisy labels \cite{pmlr-v108-li20j}. Another possible explanation is that in some cases, incorrect labels could help to "smooth" the decision boundary. This can also explain why no difference in balanced accuracy was observed only in part of our experiments. Nonetheless, more research is needed to better understand this phenomenon and its impact on self-labeling performance.
Without prior filter, models trained with smaller seed sizes accumulate erroneous labels faster in the initial phase of training. For full SL2S the fraction of wrong labels roughly states the same across the whole training. We verify this further in appendix \ref{appendix:begin_accuracy} and show that, indeed, initial model performance has no great impact on final accuracy of SL2S. 

\section{Lessons Learned}

Based on the results provided in Tab.~\ref{tab:variable_budget} we can claim that SL2S method works better for big datasets. This was expected, as for a larger stream, more samples can be accumulated with self-labeling. In the case of a smaller dataset, the performance is similar to other methods. The budget does not have a huge impact on the experiment results. Also, experiments with seed size confirm that our method could be applied for low data regimes.

As indicated by results with the nursery dataset and ablation results, a prior filter may not be the best method to address the imbalance issue in our datasets. This part of the algorithm was designed with synthetic data. In the case of real datasets, the prior class distribution has higher importance. For this reason, alternative methods should be developed for dealing with imbalance when applying self-labeling to active learning scenarios. Although we did not manage to mitigate the imbalance problem properly, solving this issue is important, and future work in this area should address this problem.

Results from Fig.~\ref{fig:initial_acc_impatct} suggest that after the budget ends, the balanced accuracy roughly stays at the same level, and changes in the test accuracy do not occur frequently. With a higher number of model updates, the performance over time could fall drastically. For this reason, we introduced solutions that limit the use of self-labeling, preventing a fall in accuracy. However, this is sub-optimal, as in this case, we would ideally want accuracy to increase over time, despite the end of the budget.
More work is needed to better address the dynamic imbalance issue or to provide a more accurate filter for wrong predictions. With these two problems solved we could give up the mechanisms that inhibit learning. It should allow for obtaining better performance, especially for bigger datasets.

\section{Conclusions}

We have proposed a new active learning method that combines simple ensemble-based sample selection and self-labeling for selective sampling. Experiments with multiple baselines show that our method offers comparable performance to other active learning algorithms for smaller datasets and better performance for bigger datasets. Further experiments also show that our method could work well when the initially labeled dataset is small or when initial model accuracy is poorly trained. 

We also show that an important aspect of self-labeling is an imbalance, as bias towards a single class in model predictions could, over time, increase dataset imbalance. Another important factor is erroneous model predictions that introduce noise into the training dataset. Based on the preliminaries and ablations presented in this work, we claim that further work should focus on these two aspects to improve the overall self-labeling performance. We cannot eliminate all errors from model predictions, however, developing better methods for filtering noisy labels or models that are more robust to label noise should allow for better utilization of self-labeling.

\section*{Acknowledgment}
This work is supported by the CEUS-UNISONO programme, which has received funding from the National Science Centre, Poland under grant agreement No. 2020/02/Y/ST6/00037.

\bibliographystyle{named}
\bibliography{bibliography}

\clearpage
\appendix

\section{Algorithm for selective sampling}
\label{appendix:algorithm_selective_sampling}

In this section, we provide a general algorithm for selective sampling, that depends on some informativeness measure $m$, that is determined by a specific active learning method. This algorithm can be easily expanded to include batch training. In such case, we introduce a buffer for labeled samples and train only where this buffer is full. After training, we empty the buffer.

\begin{algorithm}
\caption{Selective sampling}
\begin{algorithmic}[1]
\Require $\mathcal{L}$ - set of labeled data, $\mathcal{U}$ - stream of N unlabeled samples, $f_{\theta}$ - model, $B$ - budget, $m(.)$ - informativeness measure, $\alpha$ - threshold for informativeness measure

\State train $f_{\theta}$ on $\mathcal{L}$
\For{$i \in \{0, N\}$} 
    \State $\hat{y} \leftarrow  f_{\theta}(\boldsymbol{x}_i)$
    \If{ $B > 0 \land m(\hat{y}) \leq \alpha$}
        \State request label $y$ for $\boldsymbol{x}_i$
        \State $\mathcal{L} \leftarrow \mathcal{L} \cup \{ (\boldsymbol{x}_i, y) \} $
        \State train $f_{\theta}$ on $\mathcal{L}$
        \State $B \leftarrow  B - 1$
    \EndIf
\EndFor
\end{algorithmic}
\label{alg:selective_sampling}
\end{algorithm}

\section{Self-labeling Selective Sampling algorithm}
\label{appendix:detailed_algorithm}

In this section, we provide the full SL2S algorithm in listing \ref{alg:self_labeling_selective_sampling}. 
First, we perform initial ensemble training with bootstrapping (lines 1-4). During sampling (line 2), we ensure that each sampled dataset has at least a single learning example from each class. Without this procedure, some software implementations of common classifiers could fail.
Next, we sample unlabeled data from stream $\mathcal{U}$. First, we gather predictions from the models in the ensemble (lines 6-8). We check if at least half of the models in the committee returned confident and consistent predictions (lines (9-10). If the condition is met, we estimate the difference between the current estimation of the predicted class prior and prior for a perfectly balanced dataset (lines 11-13). If the current class prior does not exceed the $\frac{1}{C}$, we expand our dataset with the current sample labeled by prediction (line 14), calculate new $\lambda$ value (lines 15-20), and perform bootstrapped training (line 21) according to Algorithm \ref{alg:ensemble_training}.
When the prediction is not consistent nor confident, we check if we still have the budget (line 24). If we do, a new query is created (line 25), and the dataset is updated with a new sample (line 26) and lambdas vector with a default value for labels obtained from oracle (line 27). Lastly, we perform bootstrapped training as in the previous case. 

Bootstrapped training average time complexity is $O(L \max(M E[r], Train(f))$, where $E[r]$ depends on average model supports, the value of $\tau$, and utilization of budget, therefore for simplicity we left at as expected value. Self-labeling Selective Sampling time complexity is $O(N  \max(L Pred(f), L M E[r], L Train(f) )$, where $Pred(f)$ and $Train(f)$ are time complexities of model $f$ prediction and training respectively.

\begin{algorithm}
\caption{Self-labeling Selective Sampling}
\begin{algorithmic}[1]
\Require $\mathcal{L}$ - set of labeled data, $\mathcal{U}$ - stream of unlabeled data, $ \{ f_{\theta} \}_{L} $ - ensemble of $L$ models, $B$ - budget, $\tau$ - confident prediction threshold, $k$ - number of newest samples from dataset used to estimate prior

\For{$l \in [0, L]$}
    \State Sample dataset $D_l$ by bootstrapping $\mathcal{L}$
    \State train $f_{\theta_{l}}$ on $D_l$
\EndFor
\For{$\boldsymbol{x} \in U$}
    \For{$l \in [0, L]$}
        \State $\hat{y}_l \leftarrow f_{\theta_{l}}(\boldsymbol{x})$
    \EndFor
    \State $\hat{y} \leftarrow \max_{p_{l}(y|x)} \hat{y}_{l}$
    
    \If{ $\sum_{l} \mathds{1}_{\max_{c} p_{l}(\boldsymbol{y}_c|\boldsymbol{x}) > \tau} > \frac{L}{2} 
    \land
    \forall_{\max_c p_{l}(\boldsymbol{y}_c|\boldsymbol{x}) > \tau} \hat{y}_{l} = \hat{y}  $ }
        \State $\hat{p} \leftarrow \frac{1}{k} \sum_{i=M-k}^{M} \mathds{1}_{y_{i}=\hat{y}} $
        \State $\Delta_{p} = \hat{p} - \frac{1}{C}$
        \If{ $\Delta_{p} \leq 0$ }
            \State $\mathcal{L} \leftarrow \mathcal{L} \cup \{ (x, \hat{y}) \} $
            \State $\lambda = \frac{ \max_{l, c} p_l(\boldsymbol{y}_c|\boldsymbol{x}) }{\tau} - \mathds{1}_{B = 0}$
            \State $\boldsymbol{\lambda} \leftarrow ( \boldsymbol{\lambda}, \lambda )$
            \State Bootstrapped training($\mathcal{L}, \{ f_{\theta} \}_{L}, \boldsymbol{\lambda}) $
        \EndIf
    \Else
        \If{$B > 0$}
            \State request label $y$ for $\boldsymbol{x}$
            \State $\mathcal{L} \leftarrow \mathcal{L} \cup \{ (\boldsymbol{x}, y) \} $
            \State $\boldsymbol{\lambda} \leftarrow ( \boldsymbol{\lambda}, 1 )$
            \State $B \leftarrow  B - 1$
            \State Bootstrapped training($\mathcal{L}, \{ f_{\theta} \}_{L}, \boldsymbol{\lambda}) $
        \EndIf
    \EndIf
\EndFor

\end{algorithmic}
\label{alg:self_labeling_selective_sampling}
\end{algorithm}

\section{Data lodaing procedures}
\label{appendix:data_loading}

During data preprocessing, we split features into numerical and categorical features. For numeric features, we replace missing values with the median of a given feature and perform standard scaling by subtracting the mean and scaling to unit variance. For categorical features, we utilize only a one-hot encoder. Detailed features and their categorization is provided in Tab.~\ref{tab:data_loading}.

Some classes have a too low number of samples compared to other classes to allow for learning. For example, in the case when five classes have above 1000 learning examples each, and one class has only five samples, there is no point in keeping this class in the dataset. For this reason, we drop classes with the lowest number of samples. The list of dropped classes for each dataset is presented in Tab.~\ref{tab:classes_dropped}.

\begin{table}[ht]
    \scriptsize
    \centering
    \caption{Classes dropped from datasets}
    \begin{tabular}{c|c}
    \hline
        dataset & dropped classes \\
        \hline
        adult & - \\
        bank marketing & - \\
        firewall & reset-both \\
        chess & zero, one, three \\
        nursery & recommend \\
        poker & 7, 8, 9 \\
        mushroom & - \\
        wine & 3, 9 \\
        abalone & \pbox{8cm}{32, 20, 3, 21, 23, 22,\\ 27, 24, 1, 26, 29, 2, 25} \\
        \hline
    \end{tabular}
    \label{tab:classes_dropped}
\end{table}

The classification scores were too high for the nursery and mushroom datasets. For this reason, experimental evaluation of active learning algorithms is nearly impossible, as each algorithm could easily obtain a perfect or near-perfect score. To prevent this, we make the classification task harder by dropping the most informative features, selected based on the Person correlation coefficient computed between features and labels. The complete list of features used and target columns used in experiments are provided in Tab.~\ref{tab:data_loading} and \ref{tab:data_loading_big} for small and big datasets respectively.

\begin{table}[ht]
    \scriptsize
    \centering
    \caption{Detailed categorization of features and classes dropped from small datasets}

\begin{tabular}{c|c|c|c}
    \hline
    dataset & \pbox{2cm}{numeric features} & \pbox{2cm}{categorical features} & target feature \\
    \hline
    adult & 
    \pbox{2cm}{age, education-num, capital-gain, fnlwgt, capital-loss, hours-per-week} & 
    \pbox{2cm}{workclass, education, marital-status, occupation, relationship, race, sex, native-country} &
    earnings 
    \\
    \hline
    \pbox{1.5cm}{bank\\ marketing} & 
    \pbox{2cm}{age, duration, campaign,\\ pdays, previous} & 
    \pbox{2cm}{job, marital, education, default,\\ housing, loan, contact, month, poutcome} &
    y 
    \\
    \hline
    firewall & 
    \pbox{2cm}{Source Port, Destination Port, pkts sent, 
    NAT Source Port,  NAT Destination Port, Bytes, 
    Bytes Sent, Packets,  
    Bytes Received, Elapsed Time (sec), pkts received
    } &
    - &
    Action
    \\
    \hline
    chess & 
    0, 1, 2, 3, 4, 5 & 
    - &
    6 
    \\
    \hline
\end{tabular}
    \label{tab:data_loading}
\end{table}

\begin{table}[ht]
    \scriptsize
    \centering
    \caption{Detailed categorization of features and classes dropped from big datasets}
    
    \label{tab:data_loading_big}
\end{table}

\section{Hyperparameter optimization process}
\label{appendix:hyperparamters}

We perform hyperparameter tuning with random sampling \cite{JMLR:v13:bergstra12a}. Each method has the same search budget, i.e., the same number of runs. We sample 20 values from the predefined prediction thresholds for each method. Interval borders are provided in Tab.~\ref{tab:hyper}. For each prediction threshold, we perform three evaluations with different random seeds. We select the best hyperparameter values based on averaged accuracy from these three runs. Accuracy reported in the work was obtained with ten random seeds, different than the ones used in the hyperparameter tuning process.

\begin{table}[ht]
    \scriptsize
    \centering
    \caption{Intervals used for hyperparameters search process.}
    \begin{tabular}{c|c|c}
        \hline
        method & min value & max values \\
        \hline
        fixed uncertainty & 0.5 & 1.0 \\
        variable uncertainty & 0.5 & 1.0 \\
        classification margin & 0.0 & 0.8 \\
        vote entropy & 1.0 & 50.0 \\
        consensus entropy & 0.1 & 1.0 \\
        max disagreement & 1.0 & 20.0 \\
        min margin & 0.0 & 0.5 \\
        ours & 0.5 & 1.0 \\
        \hline
    \end{tabular}
    \label{tab:hyper}
\end{table}

\section{Results of experiments with different seed size}
\label{appendix:seed_size}

We provide full results of experiments with seed size Tab.~\ref{tab:seed_experiments}. A description of these results is provided in the main part of our paper.

\begin{table}[!ht]
    \scriptsize
    \centering
    \caption{Balanced accuracy for various seed sizes used for initial training of the model.}
    \begin{tabular}{l|ccc}
    \hline 
    & \multicolumn{3}{c}{dataset nursery} \\ 
    \hline 
     all labeled           & & 0.318±0.030 & \\
     all labeled ensemble   & & 0.276±0.013 &  \\
     \hline
    seed size & 100 & 500 & 1000 \\ 
    \hline 
 random                 &  0.341±0.017  &  0.327±0.009  &  0.325±0.012 \\
 fixed uncertainty      &  0.386±0.014  &  0.388±0.012  &  0.385±0.007 \\
 variable uncertainty   &  0.341±0.011  &  0.338±0.016  &  0.327±0.014 \\
 classification margin  &  0.382±0.012  & \textbf{  0.402±0.021  } & \textbf{  0.396±0.013 } \\
 vote entropy           &  0.354±0.010  &  0.394±0.011  &  0.393±0.014 \\
 consensus entropy      &  0.368±0.011  &  0.389±0.012  &  0.393±0.014 \\
 max disagreement       &  0.356±0.012  &  0.401±0.010  &  0.393±0.014 \\
 min margin             & \textbf{  0.394±0.019  } &  0.358±0.012  &  0.388±0.021 \\
 SL2S                   &  0.361±0.009  &  0.353±0.016  &  0.338±0.013 \\
    
    \hline 
    & \multicolumn{3}{c}{dataset mushroom} \\ 
    \hline 
     all labeled           & & 0.637±0.010 & \\
     all labeled ensemble   & & 0.636±0.010 & \\
    \hline
    seed size & 100 & 500 & 1000 \\ 
    \hline 
  random                 &  0.633±0.007  &  0.631±0.010  &  0.633±0.010 \\
 fixed uncertainty      &  0.627±0.011  &  0.629±0.010  &  0.633±0.012 \\
 variable uncertainty   &  0.622±0.013  &  0.631±0.011  &  0.634±0.010 \\
 classification margin  &  0.620±0.016  &  0.630±0.013  &  0.633±0.013 \\
 vote entropy           &  0.632±0.012  &  0.624±0.012  & \textbf{  0.635±0.011 } \\
 consensus entropy      &  0.633±0.011  &  0.629±0.013  &  0.633±0.011 \\
 max disagreement       &  0.596±0.023  &  0.624±0.012  &  0.630±0.011 \\
 min margin             &  0.628±0.009  &  0.631±0.010  &  0.634±0.010 \\
 SL2S                   & \textbf{  0.635±0.009  } & \textbf{  0.632±0.012  } &  0.632±0.012 \\
     
    \hline 
    & \multicolumn{3}{c}{dataset wine} \\ 
    \hline 
     all labeled           & & 0.524±0.027 & \\
     all labeled ensemble   & & 0.514±0.015 & \\
    \hline 
    seed size & 100 & 500 & 1000 \\ 
    \hline 
  random                 &  0.403±0.021  &  0.418±0.018  &  0.439±0.023 \\
 fixed uncertainty      &  0.395±0.017  &  0.421±0.020  &  0.441±0.017 \\
 variable uncertainty   &  0.406±0.019  &  0.423±0.019  &  0.437±0.022 \\
 classification margin  &  0.355±0.011  &  0.414±0.015  &  0.420±0.022 \\
 vote entropy           &  0.291±0.014  &  0.428±0.022  &  0.389±0.016 \\
 consensus entropy      & \textbf{  0.407±0.017  } &  0.427±0.013  &  0.458±0.014 \\
 max disagreement       &  0.316±0.026  &  0.356±0.020  &  0.389±0.016 \\
 min margin             &  0.401±0.012  &  0.435±0.022  & \textbf{  0.461±0.019 } \\
 SL2S                   &  0.405±0.011  & \textbf{  0.439±0.024  } &  0.451±0.022 \\

    \hline 
    & \multicolumn{3}{c}{dataset abalone} \\ 
    \hline 
     all labeled           & & 0.186±0.021 & \\
     all labeled ensemble   & & 0.188±0.012 & \\
    \hline 
    seed size & 100 & 500 & 1000 \\ 
    \hline 
  random                 &  0.180±0.013  &  0.180±0.011  &  0.178±0.007 \\
 fixed uncertainty      &  0.175±0.015  &  0.189±0.020  &  0.177±0.013 \\
 variable uncertainty   &  0.182±0.013  &  0.186±0.009  &  0.181±0.010 \\
 classification margin  &  0.178±0.013  &  0.180±0.014  &  0.187±0.013 \\
 vote entropy           & \textbf{  0.187±0.014  } &  0.189±0.011  &  0.185±0.015 \\
 consensus entropy      &  0.184±0.009  &  0.190±0.012  &  0.185±0.013 \\
 max disagreement       &  0.166±0.017  &  0.180±0.010  &  0.184±0.012 \\
 min margin             &  0.179±0.018  & \textbf{  0.191±0.013  } &  0.185±0.015 \\
 SL2S                   & \textbf{  0.187±0.022  } &  0.184±0.013  & \textbf{  0.190±0.010 } \\
     \hline
\end{tabular}
    \label{tab:seed_experiments}
\end{table}

\section{Impact of initial model accuracy}
\label{appendix:begin_accuracy}

\begin{figure}
    \centering
    \includegraphics[width=0.5\textwidth]{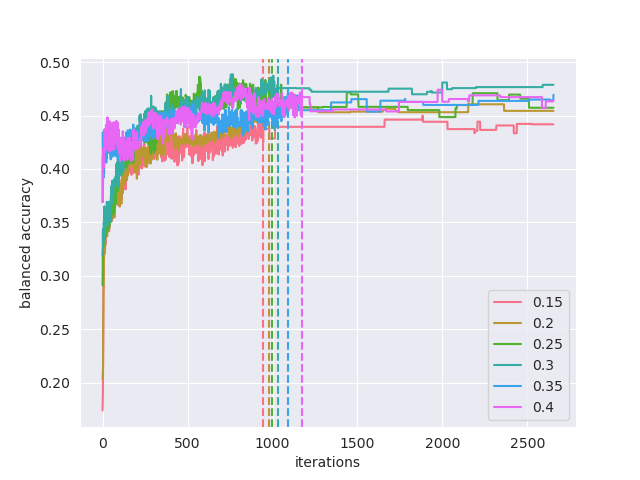}
    \caption{Impact of initial balanced accuracy on overall training results. Vertical doted lines indicate the iteration when budget ended.}
    \label{fig:initial_acc_impatct}
\end{figure}

In previous experiments, we showed that initial training set size does not have a large impact on SL2S performance. However, our primary concern are erroneous predictions produced by a weak model. For this reason, we study the relationship between the balanced accuracy of the initial model and overall experiment results. We gradually increase the seed size from 10 samples up and track test accuracy. When test accuracy exceeds the predefined value, we stop initial training and proceed to the further part of selective sampling. We evaluate models with initial accuracy equal to 0.15, 0.2, 0.25, 0.3, 0.35, and 0.4, and with a budget of 0.3. Results are plotted in Fig.~\ref{fig:initial_acc_impatct}. 

From this plot, we deduce that low accuracy at the beginning can be easily compensated by spending the budget on initial model improvement. Looking at the balanced accuracy values over time, we can see that initial values of accuracy for poorly trained models increase abruptly and the beginning of training. Higher initial budget consumption can be further exemplified by the iteration number when the budget runs out. For larger initial accuracy, the budget end is later.  
It shows that our method can be used even with a poorly trained initial model.

\end{document}